\documentclass[12pt,reqno]{amsart}
\usepackage{amsaddr}
\usepackage{amssymb,epsf}
\usepackage{amstext,amsmath,amsfonts,amscd,amsthm,graphicx}
\usepackage{upref}
\usepackage{epsfig}
\usepackage[utf8]{inputenc}
\usepackage{latexsym}
\usepackage{eucal}
\usepackage{wrapfig}
\usepackage{color}
\newfont{\fnt}{cmsy10}
\newfont{\sss}{cmti10}

\begin{document}

\title[Optimisation of complex product innovation processes based on trends]{Optimisation of complex product innovation processes based on trend models \\ with three-valued logic}

\author{Nina~Bo\v{c}kov\'{a}$^1$, Barbora~Voln\'{a}$^2$*, Mirko~Dohnal$^3$}

\address{\footnotesize $^1$Prague University of Economics and Business, Faculty of Business Administration, W. Churchilla 4, \\ 130 67 Praha 3, Czech Republic. 
\\ $^2$Silesian University in Opava, Mathematical Institute in Opava, Na Rybn\'{i}\v{c}ku 1, \\ 746 01 Opava, Czech Republic.
\\ $^3$Brno University of Technology, Faculty of Business and Management, Kolejn\'{i} 2906/4, \\ 61200 Brno, Czech Republic.
\\ *Corresponding author}
\email{nina.bockova@vse.cz}
\email{Barbora.Volna@math.slu.cz}
\email{dohnal@vutbr.cz}


\begin{abstract}
This paper investigates complex product-innovation processes using models grounded in a set of heuristics. Each heuristic is expressed through simple trends -- increasing, decreasing, or constant -- which serve as minimally information-intensive quantifiers, avoiding reliance on numerical values or rough sets. A solution to a trend model is defined as a set of scenarios with possible transitions between them, represented by a transition graph. Any possible future or past behaviour of the system under study can thus be depicted by a path within this graph.
\end{abstract}

\maketitle

\section{Introduction}

Product innovations (PI) are creative, multidimensional, complex, and often poorly understood processes of an interdisciplinary nature, studied for instance in \cite{alegre_lapiedra_chiva}, \cite{arundel_smith}, and \cite{downs_velamuri}. A key difficulty in nearly all realistic PI models is the shortage of available information. Statistical approaches rely on the law of large numbers, which requires a substantial number of observations. Consequently, unique creative tasks can only rarely be analysed accurately using traditional statistical methods \cite{sandroni_urgun}. This limitation has motivated the increasing use of new formal tools, such as fuzzy and rough sets \cite{nasiri_huang}, \cite{vesely_klockner_dohnal}.

Against this background, it is important to clarify the role of different quantifiers in modelling PI processes. There are several types of quantifiers. The most frequently used are numerical values (see \cite{atkinson_kress_szechtman}, \cite{vitanov_vitanov}). Other examples include fuzzy sets and verbal quantifiers. Quantifiers differ in their degree of information intensity. The least information-intensive are trends -- decreasing, constant, or increasing. If a trend cannot be quantified or predicted, then nothing can be measured, observed, or ultimately decided. This implies that quantitatively oriented PI decision-making procedures cannot be applied \cite{royes_bastos}, \cite{wright_goodwin}.

Deep knowledge items are laws that reflect undisputed elements of relevant theories. For instance, Einstein’s theory of relativity represents a deep knowledge item. Such items are typically available in mathematical form, for example as a set of differential or algebraic equations. Unfortunately, only a limited number of deep knowledge items can be identified in the context of PI.

A shallow knowledge item is a heuristic or the outcome of statistical analysis based on passive observations. It is evident that PI does not permit active experimentation when large-scale production facilities are under study. Moreover, shallow knowledge items typically involve numerous exceptions.

In line with this approach, we present and extend trend-based modelling as an analytical tool for optimising complex product innovation processes. The remainder of the paper is organised as follows. The first part of the study outlines trend-based modelling, including the trend correlation matrix, trend consistency, the integration of objective and subjective knowledge items, and transition graphs. The second part, a case study, formulates and solves the trend-based PI model and demonstrates its optimisation.

\section{Trend-Based Modelling as an Analytical Tool}

In this section, we address the interpretation of the correlation matrix from the perspective of trend-based modelling and the consistency of the trend model. 
We also integrate objective and subjective knowledge items, formalise the definition of a trend model, and present the transition graph as a suitable graphical representation of the solution to a trend-based model.

\subsection{Trend Interpretation of Correlation Matrix}

Various types of correlation matrices exist \cite{bun_bouchaud_potters}. They are widely used and therefore well established \cite{paul_aue,shevlyakov_oja}. Moreover, correlation matrices have been applied in a broad range of contexts \cite{frigessi_loland_pievatolo_ruggeri}.

It is possible to construct a trend model from the information contained in a given deterministic correlation matrix; see the rules described in \eqref{correlation_matrix_rules}. Let $C \in \mathbb{R}^{n \times n}$ be a correlation matrix and let $c_{ij}$ denote the correlation coefficient between variables $X_i$ and $X_j$, i.e., $C = (c_{ij})$. The trend-based correlation matrix can then be expressed using the following notation, see also \cite{dohnal,doubravsky_dohnal}:
\begin{equation}
\label{correlation_matrix_rules}
\begin{array}{lll}
\text{If } c_{ij} > 0 & \text{then } \operatorname{INC}(X_i, X_j) & \text{indicates a positive correlation},\\
\text{If } c_{ij} < 0 & \text{then } \operatorname{DEC}(X_i, X_j) & \text{indicates a negative correlation}.
\end{array}
\end{equation}
Note that coefficients with $c_{ij} = 0$ are omitted, as they indicate no relationship and therefore are not included in the trend-based model.

\subsection{Trend Model Consistency}

The statistical nature of a classical numerical correlation matrix can cause problems when a trend model is derived from it using the rules in \eqref{correlation_matrix_rules}. In practice, the correlation matrix is almost always statistically inconsistent. The deterministic interpretation given by \eqref{correlation_matrix_rules} may therefore generate a trend model that has no solution.

A trivial brute-force (exhaustive) search can be used to solve simple trend models. For example, the following simple case:
\begin{equation}
\label{example_res_mod}
\begin{array}{l}
\text{INC X Y}\\
\text{DEC X Z}\\
\text{INC Y Z}
\end{array}
\end{equation}
yields only a steady-state solution; see Table~\ref{tab:steady_state_solution}.

\begin{table}[ht]
\centering
\begin{tabular}{cccc}
\hline
  & $X$     & $Y$     & $Z$ \\
\hline
1 & $+~0~*$ & $+~0~*$ & $+~0~*$ \\
\hline \hline
\end{tabular}
\vspace{0.1cm}
\caption{Steady-state solution of model \eqref{example_res_mod}}
\label{tab:steady_state_solution}
\end{table}

In Table~\ref{tab:steady_state_solution}, the first sign ($+$) of any triplet indicates that the variable is positive; the second symbol ($0$) means that the first time derivative equals zero; and the asterisk ($*$) denotes that the second time derivative is unrestricted.

In the case of a unique scenario corresponding to the steady state (see Table~\ref{tab:steady_state_solution}), all three first time derivatives, denoted by $DX$, $DY$, and $DZ$, are equal to zero. This clearly indicates that the model is restrictive and that no dynamic behaviour is possible. However, if the very nature of the variables $X$, $Y$, and $Z$ suggests that dynamic scenarios should exist, then the model \eqref{example_res_mod} is over-restrictive and should be rectified.

A rectification of model \eqref{example_res_mod} can be achieved by removing one or more rows, that is, specific statements or relations. If the last row of \eqref{example_res_mod} is removed, the three scenarios described in Table~\ref{tab:12rows_solution} are obtained. If, on the other hand, the second row of model~\eqref{example_res_mod} is removed, the modified model yields the set of scenarios shown in Table~\ref{tab:13rows_solution}. As we can see, each modification of model~\eqref{example_res_mod} produces a different set of scenarios (see Tables~\ref{tab:12rows_solution} and~\ref{tab:13rows_solution}).

\begin{table}[ht]
\centering
\begin{tabular}{cccc}
\hline
  & $X$     & $Y$     & $Z$ \\
\hline
1 & $++*$   & $++*$   & $+-*$ \\
2 & $+~0~*$ & $+~0~*$ & $+~0~*$ \\
3 & $+-*$   & $+-*$   & $++*$ \\
\hline \hline
\end{tabular}
\vspace{0.1cm}
\caption{Solution of model \eqref{example_res_mod} without the third row}
\label{tab:12rows_solution}
\end{table}

\begin{table}[ht]
\centering
\begin{tabular}{cccc}
\hline
  & $X$     & $Y$     & $Z$ \\
\hline
1 & $++*$   & $++*$   & $++*$ \\
2 & $+~0~*$ & $+~0~*$ & $+~0~*$ \\
3 & $+-*$   & $+-*$   & $+-*$ \\
\hline \hline
\end{tabular}
\vspace{0.1cm}
\caption{Solution of model \eqref{example_res_mod} without the second row}
\label{tab:13rows_solution}
\end{table}

Naturally, there exist models that are not restrictive; for example, model~\eqref{example_not-res_mod}. Model~\eqref{example_not-res_mod} yields a set of three scenarios, as shown in Table~\ref{tab:not-res_mod_solution}.
\begin{equation}
\label{example_not-res_mod}
\begin{array}{l}
\text{DEC X Y}\\
\text{INC Y Z}\\
\text{DEC X Z}
\end{array}
\end{equation}

\begin{table}[ht]
\centering
\begin{tabular}{cccc}
\hline
  & $X$     & $Y$     & $Z$ \\
\hline
1 & $++*$   & $+-*$   & $+-*$ \\
2 & $+~0~*$ & $+~0~*$ & $+~0~*$ \\
3 & $+-*$   & $++*$   & $++*$ \\
\hline \hline
\end{tabular}
\vspace{0.1cm}
\caption{Solution of model \eqref{example_not-res_mod}}
\label{tab:not-res_mod_solution}
\end{table}

Generally, a \textit{trend model} $M$ contains $w \in \mathbb{N}$ pairwise relations (rows). For example, model~\eqref{example_not-res_mod} includes three such relations, i.e., $w = 3$. We can formally write $M = (P_1, P_2, \dots, P_w)$, where each $P_k$ for $k = 1, 2, \dots, w$ represents a pairwise relation (row) of the model, see e.g. \cite{dohnal,doubravsky_dohnal}.

Additionally, we define a \textit{restrictive trend model}, denoted by $M_R$, as a model that has the steady-state scenario as its single solution. For instance, model~\eqref{example_res_mod} has a unique steady-state solution (see Table~\ref{tab:steady_state_solution}) and is therefore restrictive.

Let $0 < v < w$. A modification of a restrictive trend model $M_R = (P_1, \dots, P_w)$ intended to remove its restrictiveness (i.e., restrictive relations) consists in removing $w - v$ relations (rows) from the original model $M_R$ to establish a new trend model $M_N = (P_{N_1}, \dots, P_{N_v})$. 
In summary,
\begin{equation}
\label{modification}
M_R = (P_1, \dots, P_w) \rightarrow M_N = (P_{N_1}, \dots, P_{N_v}), \quad w > v.
\end{equation}

The modification \eqref{modification} is not unique. For example, the model~\eqref{example_res_mod} is a restrictive trend model and can be modified either by removing the third row, yielding a non-restrictive model with the solution shown in Table~\ref{tab:12rows_solution}, or by removing the second row, resulting in a non-restrictive model with the solution given in Table~\ref{tab:13rows_solution}.

Therefore, the elimination of restrictiveness can be formulated as an optimisation problem. Different optimisation objectives can be defined, for example:
\begin{itemize}
\item[O1:] Minimise the number of removed relations, i.e., $w - v$.
\item[O2:] Minimise $\sum |c_{ij}|$, where $c_{ij} \in R$ and $R$ is the set of removed pairwise trend relations of model $M$ satisfying the modification \eqref{modification}.
\end{itemize}

The optimisation objective O2 eliminates those pairwise relations that have low values of the corresponding correlation coefficients $c_{ij}$. Additionally, the identification of model $M_R$ using the objective function O2 becomes a complex process when the number of relations $w$ is large.

However, a potential user of the model $M_R$ usually has their own elimination preferences and removes relations according to common sense. Therefore, an ad hoc, common-sense-based model, denoted by $M_{CS}$ and derived from \eqref{modification}, can be expressed as:
\begin{equation}
\label{common-sense_model}
M_R = (P_1, \dots, P_w) \rightarrow M_{CS} = (P_{N_1}, \dots, P_{N_v}), \quad w > v.
\end{equation}

Note that two models, $M_A$ and $M_B$, are considered equivalent if their sets of scenarios, $S_A$ and $S_B$, are identical. However, equivalent models are not examined in this paper.

\subsection{Integration of Objective and Subjective Knowledge Items}

\begin{figure}[ht]
  \centering
  \includegraphics[height=3.5cm]{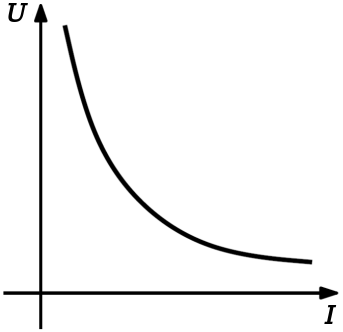}
  \caption{Illustration of the example of a subjective knowledge item: variable $U$ decreases with increasing $I$, while the rate of decrease gradually slows down (a convex decreasing trend)}
  \label{fig:decreasing_convex}
\end{figure}

Correlation matrices represent an objective source of information, derived either from active experimental results (e.g., in chemistry or physics) or from passive observations (e.g., in economics or sociology). However, subjective knowledge items are usually available as well, for instance: if $I$ is increasing, then $U$ is decreasing, and the decrease in $U$ gradually slows down. The graphical representation of this subjective knowledge item is illustrated in~Figure \ref{fig:decreasing_convex}.

Such subjective knowledge items can be made more formal if the second trend derivatives, i.e., trends of trends, are available. This means that the trend relation corresponding to our example is expressed as
\begin{equation}
\frac{d^2 U}{dI^2} > 0 .
\end{equation}

\begin{figure}[ht]
  \centering
  \includegraphics[height=3.55cm]{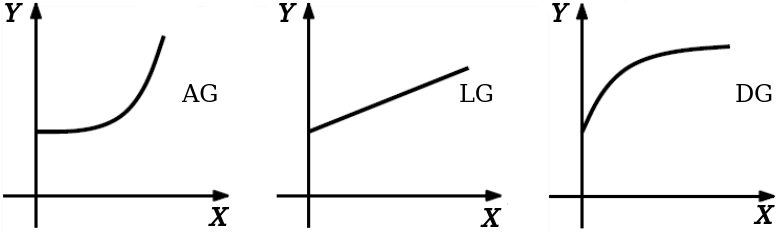}
  \caption{Illustration of accelerating growth, linear growth, and decelerating growth}
  \label{fig:AG_LG_DG}
\end{figure}

\begin{figure}[ht]
  \centering
  \includegraphics[height=3.55cm]{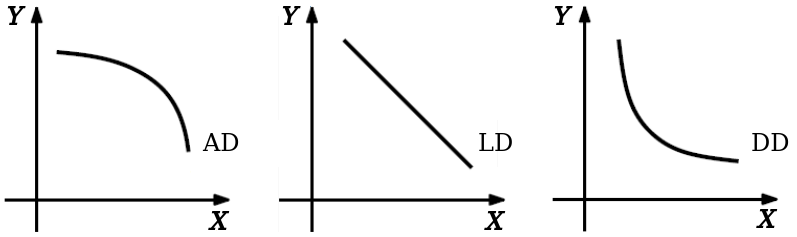}
  \caption{Illustration of accelerating decrease, linear decrease, and decelerating decrease}
  \label{fig:AD_LD_DD}
\end{figure}

Examples of positive and negative relations between positive-valued variables are shown in Figures~\ref{fig:AG_LG_DG} and~\ref{fig:AD_LD_DD}, respectively. 
The notations AG, LG, DG, AD, LD, and DD represent accelerating growth, linear growth, decelerating growth, accelerating decrease, linear decrease, and decelerating decrease, respectively.

In this paper, these relations are formalised using the identification notation introduced in Figures~\ref{fig:AG_LG_DG} and~\ref{fig:AD_LD_DD}, and we follow up on e.g. \cite{dohnal,doubravsky_dohnal}. For instance, the relation illustrated in Figure~\ref{fig:decreasing_convex} is written as DD~$I$~$U$.

All pairwise relations described in Figures~\ref{fig:AG_LG_DG} and~\ref{fig:AD_LD_DD} are referred to as \textit{trend relations}. This means that nothing is known quantitatively, e.g., no numerical values are specified.
\newpage For example, the relation DG indicates that:
\begin{itemize}
\item the corresponding function $Y(X)$ is increasing, i.e., the first derivative $\frac{dY}{dX} > 0$;
\item this increase becomes progressively slower, i.e., the second derivative $\frac{d^2Y}{dX^2} < 0$;
\item the variables $X$ and $Y$ are positive-valued, and $Y$ is positive even for $X = 0$.
\end{itemize}

\subsection{Formal Definition of a Trend Model}

Formally, an \textit{$n$-dimensional trend model} $M$ is represented by a finite set of $w$ pairwise relations \cite{dohnal,doubravsky_dohnal}
\begin{equation}
\label{trend_model}
M = \{P_k(X_i, X_j) \mid k = 1,2 \dots w; (i, j) \in I,I \subseteq \{ (p, q) \mid p, q \in \{1, 2, \dots, n\}, p \neq q \}\},
\end{equation}
where each $P_k$ denotes a unique pairwise trend relation between variables $X_i$ and $X_j$. These relations may be derived either from a correlation matrix according to the rules in \eqref{correlation_matrix_rules}, or from subjective knowledge items described by the identification notation in Figures~\ref{fig:AG_LG_DG} and~\ref{fig:AD_LD_DD}. Note that the model $M$ does not necessarily contain all possible pairwise relations between variables $X_i$ and $X_j$; it therefore represents a subset of all possible combinations.

Considering all pairwise relations and using dedicated software, we can find the solutions to the trend model~\eqref{trend_model}. In fact, solving a trend model means solving the combinatorial task that integrates all available knowledge, both objective (derived from the correlation matrix) and subjective. 
The solution is represented by a set of $m$ scenarios \cite{dohnal,doubravsky_dohnal}:
\begin{equation}
\label{set_of_scenarios}
S(m,n) = \{S_r(X_i, DX_i, DDX_i) \mid r = 1, 2, \dots, m;\, i = 1, 2, \dots, n\},
\end{equation}
where $X_i$ denotes the sign ($+$, $-$, or $0$) of the $i$-th variable, $DX_i$ the sign of its first time derivative, and $DDX_i$ the sign of its second time derivative.

\subsection{Transitional Graphs}

The set $S$ of $m$ scenarios defined in \eqref{set_of_scenarios} is not the only result of the trend model. It is also possible to generate $T$ \textit{time transitions} among the scenarios in $S$. A complete set of all possible one-dimensional transitions (for a single variable $X$, i.e., $n = 1$) is given in Tables~\ref{tab:transitions_positive},~\ref{tab:transitions_negative}, and~\ref{tab:transitions_zero}, see also \cite{dohnal,doubravsky_dohnal}.

\begin{table}[ht]
\centering
\begin{tabular}{c l l l l l l l l}
\hline
No. & From & To & Alt. 1 & Alt. 2 & Alt. 3 & Alt. 4 & Alt. 5 & Alt. 6 \\	
\hline					
1 & $+++$   & $++\,0$ &         &         & & & & \\						
2 & $++\,0$ & $+++$   & $++-$   &         & & & & \\
3 & $++-$   & $++\,0$ & $+~0~-$ & $+~0~0$ & & & & \\				
4 & $+~0~+$ & $+++$   &         &         & & & & \\						
5 & $+~0~0$ & $+++$   & $+--$   &         & & & & \\					
6 & $+~0~-$ & $+--$   &         &         & & & & \\						
7 & $+-+$   & $+-\,0$ & $+~0~+$ & $+~0~0$ & $0\,-+$ & $0~0~+$ & $0~0~0$ & $\,0-0$ \\
8 & $+-\,0$ & $+-+$   & $+--$   & $\,0-0$   & & & & \\				
9 & $+--$   & $+-\,0$ & $0~--$  & $\,0-0$   & & & & \\				
\hline \hline
\end{tabular}
\vspace{0.1cm}
\caption{Set of possible one-dimensional transitions for an initially positive-valued variable}
\label{tab:transitions_positive}
\end{table}

\begin{table}[ht]
\centering
\begin{tabular}{c l l l l l l l l}
\hline
No. & From & To & Alt. 1 & Alt. 2 & Alt. 3 & Alt. 4 & Alt. 5 & Alt. 6 \\	
\hline					
1 & $-++$   & $-+\,0$ & $0\,++$ & $\,0+0$ & & & & \\						
2 & $-+\,0$ & $-+-$   & $-++$   & $\,0+0$ & & & & \\
3 & $-+-$   & $-+\,0$ & $-~0~-$ & $-~0~0$ & $0\,+-$ & $0~0~-$ & $0~0~0$ & $\,0+0$\\ 4 & $-~0~+$ & $-++$   &         &         & & & & \\						
5 & $-~0~0$ & $-++$   & $---$   &         & & & & \\					
6 & $-~0~-$ & $---$   &         &         & & & & \\						
7 & $--+$   & $--\,0$ & $-~0~+$ & $-~0~0$ & & & & \\
8 & $--\,0$ & $---$   & $--+$   &         & & & & \\				
9 & $---$   & $--\,0$ &         &         & & & & \\				
\hline \hline
\end{tabular}
\vspace{0.1cm}
\caption{Set of possible one-dimensional transitions for an initially negative-valued variable}
\label{tab:transitions_negative}
\end{table}

\begin{table}[ht]
\centering
\begin{tabular}{c l l l l}
\hline
No. & From & To & Alt. 1 & Alt. 2 \\	
\hline					
1 & $0++$   & $++\,0$ & $++-$ & $+++$ \\						
2 & $0+\,0$ & $++\,0$ & $++-$ & $+++$ \\
3 & $0+-$   & $++-$   &       & \\
4 & $0~0~+$ & $+++$   &       & \\						
5 & $0~0~0$ & $+++$   & $---$ & \\					
6 & $0~0~-$ & $---$   &       & \\						
7 & $0-+$   & $--+$   &       & \\
8 & $0-\,0$ & $--\,0$ & $--+$ & $---$ \\				
9 & $0--$   & $--\,0$ & $--+$ & $---$ \\				
\hline \hline
\end{tabular}
\vspace{0.1cm}
\caption{Set of possible one-dimensional transitions for an initially zero-valued variable}
\label{tab:transitions_zero}
\end{table}

For example, the third line in Table~\ref{tab:transitions_positive} indicates that it is possible to transform the triplet $++-$ into the triplet $++\,0$. There are two additional possible transitions: $+~0~-$ and $+~0~0$.

The resulting trend behaviour of the variable $X$ should be smooth, without any jumps, steps, or kinks in time, which corresponds to the possible one-dimensional transitions shown in Tables~\ref{tab:transitions_positive},~\ref{tab:transitions_negative},~and~\ref{tab:transitions_zero}. 

\begin{figure}[ht]
  \centering
  \includegraphics[height=5.5cm]{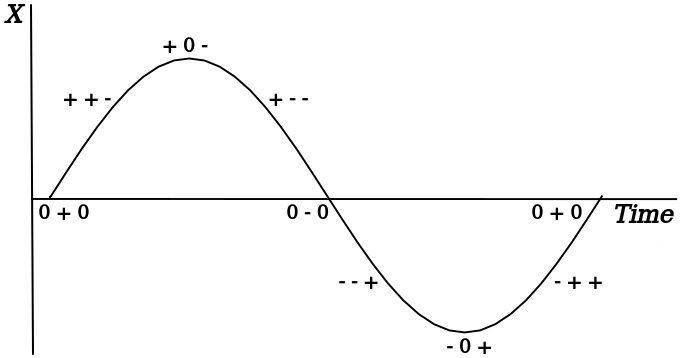}
  \caption{Illustration of one-dimensional oscillation of the variable $X$}
  \label{fig:oscillation}
\end{figure}

An example of such smooth trend behaviour is a one-dimensional oscillation, as illustrated in Figure~\ref{fig:oscillation}, see also \cite{dohnal,doubravsky_dohnal}. In this case, the triplets $0+0$ and $0-0$ correspond to points where the variable $X$ simultaneously crosses the time axis and changes its curvature -- that is, inflection points where the trend switches from decelerating to accelerating (or vice versa). The triplets $+0-$ and $-0+$ represent the local maximum and minimum of $X$, respectively. The triplets $++-$ and $+--$ represent decelerating growth and decrease of the positive-valued variable $X$, while the triplets $--+$ and $-++$ represent accelerating decrease and growth of the negative-valued variable $X$.

The graphical representation of the trend model time transitions is referred to as a \textit{transitional graph}. A transitional graph $T$ is a directed (oriented) graph whose nodes correspond to the set of scenarios $S$ defined in \eqref{set_of_scenarios}, and whose directed arcs represent the transitions following from the trend model. A transitional graph represents all possible non-steady-state behaviours of the set of cooperating and/or competing components of the trend model (i.e., the variables $X$ of the model), assuming that the model is correct. 
Therefore, any forecast corresponds to selecting a path through the transitional graph.

\begin{figure}[ht]
  \centering
  \includegraphics[height=5.5cm]{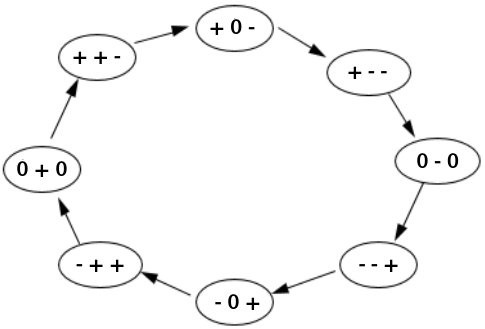}
  \caption{Transitional graph corresponding to the one-dimensional oscillation of the variable $X$}
  \label{fig:transitional_graph_cycle}
\end{figure}

As an example, a transitional graph corresponding to the one-dimensional oscillation of the variable $X$ (see Figure~\ref{fig:oscillation}) is shown in Figure~\ref{fig:transitional_graph_cycle}.

\section{Case Study}

Trend models of PI represent a broad spectrum of innovation tasks. The presented case study provides a trend-based interpretation of knowledge transfers within a network of daughter companies belonging to an international enterprise. The profitability and competitiveness of multinational enterprises are driven by their ability to accumulate and exploit gradually acquired technological knowledge~\cite{beugelsdijk_jindra}. This process is inherently multidimensional. Therefore, the considered variables are shown in Table~\ref{tab:case_study_variables}.

\begin{table}[ht]
\centering
\begin{tabular}{l l l}
\hline
Variable & Desirable trend & Abbr. \\
\hline
Product innovation                  & Increase & PI \\
Subsidiary decision-making autonomy & Increase & SD \\
External embeddedness               & Decrease & EE \\
Subsidiary size                     & Neutral  & SS \\
Subsidiary age                      & Neutral  & SA \\
Share of highly educated employees  & Increase & SE \\
Knowledge sourcing                  & Neutral  & KS \\
\hline \hline
\end{tabular}
\vspace{0.1cm}
\caption{Variables of the model in the case study}
\label{tab:case_study_variables}
\end{table}

A team of postdoctoral researchers and PhD students was asked to develop a trend model based on the set of knowledge items provided in~\cite{beugelsdijk_jindra}. However, human common-sense reasoning based on experience is often the only additional source of information or knowledge items. Consequently, the knowledge base given in~\cite{beugelsdijk_jindra} was heavily modified. The team identified, for example, the feature
\begin{quote}
\itshape
No tree can grow to Heaven.
\end{quote}
represented by the shape DG in Figure~\ref{fig:AG_LG_DG}. The following model (see Table~\ref{tab:case_study_model_rows}) was considered the best variant:
\begin{table}[ht]
\centering
\begin{tabular}{c l}
\hline
No. &  Pairwise relation\\
\hline
1 & DEC SA PI \\ 
2 & DEC EE PI \\ 
3 & DEC EE SD \\
4 & DG SS SA \\
5 & DEC KS PI \\
6 & DEC SA SS \\
7 & INC SE EE \\
8 & INC SE SS \\
\hline \hline
\end{tabular}
\vspace{0.1cm}
\caption{Model rows in the case study}
\label{tab:case_study_model_rows}
\end{table}

In Table~\ref{tab:case_study_scenarios}, seven scenarios generated by the trend model presented in Table~\ref{tab:case_study_model_rows} are shown.
\begin{table}[ht]
\centering
\begin{tabular}{clllllll}
\hline
  &   PI    &   SD    &   EE    &   SS    &   SA    &   SE    &   KS     \\
\hline
1 & $+++$   & $+++$   & $+++$   & $+++$   & $+++$   & $+--$   & $+++$    \\
2 & $++-$   & $++-$   & $++-$   & $++-$   & $++-$   & $+-+$   & $++-$    \\
3 & $+~0~+$ & $+~0~+$ & $+~0~+$ & $+~0~+$ & $+~0~+$ & $+~0~-$ & $+~0~+$  \\
4 & $+~0~0$ & $+~0~0$ & $+~0~0$ & $+~0~0$ & $+~0~0$ & $+~0~0$ & $+~0~0$  \\
5 & $+~0~-$ & $+~0~-$ & $+~0~-$ & $+~0~-$ & $+~0~-$ & $+~0~+$ & $+~0~-$  \\
6 & $+-+$   & $+-+$   & $+-+$   & $+-+$   & $+-+$   & $++-$   & $+-+$    \\
7 & $+--$   & $+--$   & $+--$   & $+--$   & $+--$   & $+++$   & $+--$    \\
\hline \hline
\end{tabular}
\vspace{0.1cm}
\caption{Solution of the model used in the case study}
\label{tab:case_study_scenarios}
\end{table}

We can observe that scenario no.~4 represents a steady state (unchanging in time); see Table~\ref{tab:case_study_scenarios}.

Four objective functions are studied; see Table~\ref{tab:case_study_objective_functions}. These functions correspond to the model variables with desirable trends; see Table~\ref{tab:case_study_variables}.
\begin{table}[ht]
\centering
\begin{tabular}{l l l}
\hline
Objective function & Desirable trend & Abbr. \\
\hline
Product innovation                  & Increase & PI \\
Subsidiary decision-making autonomy & Increase & SD \\
External embeddedness               & Decrease & EE \\
Share of highly educated employees  & Increase & SE \\
\hline \hline
\end{tabular}
\vspace{0.1cm}
\caption{Objective functions in the case study}
\label{tab:case_study_objective_functions}
\end{table}

An increase of a trend variable is represented by the triplet ($++*$), where $*$ denotes any trend value. The triplet ($+++$) indicates that the increase is fast and accelerating; see AG in Figure~\ref{fig:AG_LG_DG}. For this reason, the triplet ($+++$) corresponding to AD is considered better than ($++-$) corresponding to DG; see Figure~\ref{fig:AG_LG_DG}. Unfortunately, there is no scenario that satisfies all desired trends; see Tables~\ref{tab:case_study_scenarios} and~\ref{tab:case_study_objective_functions}.

The set of time transitions in this case study is listed in Table~\ref{tab:case_study_transitions} and contains eight items.
\begin{table}[ht]
\centering
\begin{tabular}{c c}
\hline
From scenario no. & To scenario no. \\	
\hline					
2 & 4 \\  
2 & 5 \\
3 & 1 \\
4 & 1 \\
4 & 7 \\
5 & 7 \\ 
6 & 3 \\
6 & 4 \\				
\hline \hline
\end{tabular}
\vspace{0.1cm}
\caption{All possible transitions in the case study}
\label{tab:case_study_transitions}
\end{table}

\begin{figure}[ht]
  \centering
  \includegraphics[height=5cm]{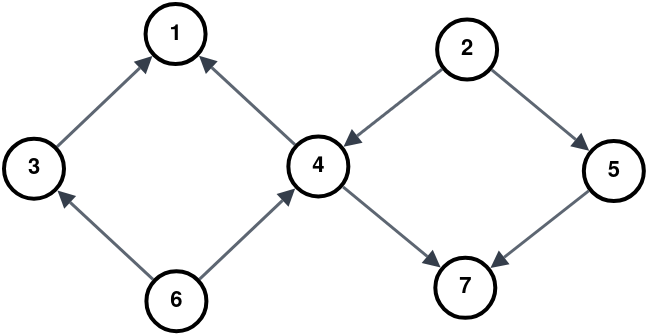}
  \caption{Transitional graph in the case study}
  \label{fig:case_study_transitional_graph}
\end{figure}

The transitional graph corresponding to the transitions listed in Table~\ref{tab:case_study_transitions} is depicted in Figure~\ref{fig:case_study_transitional_graph}.

Scenarios no.~1 and~7 (see Table~\ref{tab:case_study_scenarios}) are terminal. This means that there are no transitions enabling one to leave these scenarios. Such a situation would be advantageous if these scenarios represented desirable trends. Let us examine these two scenarios; see Table~\ref{tab:case_study_terminal_scenarios}.

\begin{table}[ht]
\centering
\begin{tabular}{cllll}
\hline
  &   PI    &   SD    &   EE    &   SE    \\
\hline
1 & $+++$   & $+++$   & $+++$   & $+--$   \\
7 & $+--$   & $+--$   & $+--$   & $+++$   \\
\hline \hline
\end{tabular}
\vspace{0.1cm}
\caption{Terminal scenarios identified in the case study}
\label{tab:case_study_terminal_scenarios}
\end{table}

If the objective functions EE and SE are sacrificed, then the first scenario satisfies the remaining required trends of the objective functions with accelerating growth. On the other hand, if we sacrifice the objective functions PI and SD, then EE and SE exhibit accelerating growth and accelerating decrease, respectively. Accelerating growth or decrease is more desirable than decelerating behaviour.

\newpage
\section{Conclusion}

At present, the techniques employed for the analysis of complex innovation tasks are of analytical or statistical natures. However, these precise mathematical tools do not always contribute as much as expected to a full understanding of the systems under study. It is, therefore, no paradox that information non-intensive methods of analysis often achieve more accurate and realistic results when the system being modelled is highly complex or little known.

The main advantages of the proposed trend method are that:
\begin{itemize}
\item No numerical values of constants or parameters are needed, and the set of trend solutions is a superset of all meaningful solutions.
\item A complete list of all futures/histories is obtained.
\item The dynamics of economic models is taken into account.
\item The results are easy to understand without knowledge of sophisticated mathematical tools.
\end{itemize}

In this paper, we provide an optimized trend-based model of a complex product innovation process. We identified a set of seven possible scenarios in the model with seven variables and four main objective functions. The transitional graph shows two terminal scenarios, both of which represent compromise options in the sense that only two of the four model objective functions satisfy the required trends (increasing/decreasing) and do so with accelerating dynamics.

In summary, trend-based modelling characterised by solutions containing the set of all possible scenarios and the set of all possible transitions between them brings new results in the field of complex product innovation processes.

\section*{Acknowledgements}
The research was supported by Prague University of Economics and Business (Faculty of Business Administration), Silesian University in Opava (Mathematical Institute in Opava) and Brno University of Technology (Faculty of Business and Management), Czech Republic.

\section*{Statement}
During the preparation of this work the authors used ChatGPT (OpenAI) in order to improve English. After using this tool, the authors reviewed and edited the content as needed and take full responsibility for the content of the publication.

\newpage
\bibliographystyle{plain}
\bibliography{references}

\end{document}